\pdfoutput=1

\documentclass[11pt]{article}

\usepackage[]{acl}

\usepackage{times}
\usepackage{latexsym}

\usepackage[T1]{fontenc}

\usepackage[utf8]{inputenc}

\usepackage{microtype}

\usepackage{inconsolata}
\usepackage{graphicx} 
\usepackage{booktabs}
\usepackage{tabularx}
\usepackage{amsmath} 

\title{Beyond Text: Unveiling Multimodal Proficiency of Large Language Models with MultiAPI Benchmark}

\author{Xiao Liu \qquad Jianfeng Lin \qquad Jiawei Zhang \\
        IFM Lab, University of California, Davis\\
        \texttt{xiao@ifmlab.org, jfglin@ucdavis.edu, jiawei@ifmlab.org}}

\begin{document}
\maketitle

\begin{abstract}
    The proliferation of Large Language Models like ChatGPT has significantly advanced language understanding and generation, impacting a broad spectrum of applications. However, these models predominantly excel in text-based tasks, overlooking the complexity of real-world multimodal information. This study introduces \textbf{MultiAPI}, a pioneering comprehensive large-scale API benchmark dataset aimed at expanding LLMs' proficiency in multimodal contexts. Developed collaboratively through ChatGPT, \textbf{MultiAPI} consists of 235 diverse API calls and 2,038 contextual prompts, offering a unique platform evaluation of tool-augmented LLMs handling multimodal tasks. Through comprehensive experiments, our findings reveal that while LLMs demonstrate proficiency in API call decision-making, they face challenges in domain identification, function selection, and argument generation. What's more, we surprisingly notice that auxiliary context can actually impair the performance.  An in-depth error analysis paves the way for a new paradigm to address these challenges, suggesting a potential direction for future LLM research.\footnote{Our data and experiment code will be available at: https://github.com/HaroldLiuJ/MultiAPI}

\end{abstract}

\section{Introduction}


Large Language Models (LLMs), such as ChatGPT, have emerged as powerful tools in understanding and generating human language \cite{li2023reinforcement, touvron2023llama, openai2023gpt4}, playing a pivotal role in diverse open-domain tasks and leaving a significant impact on both industry and academia \cite{bubeck2023sparks, yao2023react, touvron2023llama, laskar-etal-2023-systematic}. However, their performance is often confined to the text-based domains and tasks they were trained on, overlooking the multimodal and dynamic nature of real-world information. As people increasingly rely on LLMs to address their daily challenges, the imperative to enhance the task-handling capabilities of these models grows ever more pressing. In addition to addressing many of people's emerging needs in the real world, enhancing LLMs with multimodal problem-solving skills could be a significant step towards the realization of AGI in an idealized future \cite{bubeck2023sparks}.

Reflecting this demand and vision, recent studies have embarked on two primary approaches to integrate multimodal processing capabilities into existing LLMs \cite{li2023multimodal}: 1) Joint training or finetuning LLMs with components for multimodal encoding and generation \cite{wu2023nextgpt, maaz2023videochatgpt, zhang2023videollama}; 2) Introducing auxiliary API tools via natural language interfaces \cite{patil2023gorilla, shen2023hugginggpt, qin2023toolllm}, positioning LLMs as the central decision-making entity determining the appropriate tools to employ for the inquiry. Joint training of multimodal LLMs, despite creating more unified models, faces challenges with computational demands and potential loss of the generalization ability \cite{bubeck2023sparks}. On the other hand, evolving API functions, which are modularly designed, allow LLMs to adapt to new tasks by simply altering the API configuration.


 
Despite the significant potential and flexibility the tool-augmented LLMs express on multimodal tasks, their quantitative performance of multimodal tasks when integrated with API tools still remains insufficiently examined. Recent studies are very inadequate and merely focus on and gleaning insights from open-domain tasks such as mathematical computations, database searches, and graph reasoning  \cite{li2023apibank, zhuang2023toolqa, qiu2023multisum}. This gap in leveraging  API tools to achieve multimodal tasks can be attributed to two primary obstacles: 1) the unavailability of high-quality API-prompt datasets, and 2) the absence of established metrics specifically designed to evaluate the efficacy of LLMs in multimodal tasks.

In this paper, we address the aforementioned challenges by constructing a large-scale API instruction-function dataset that provides API functions and evaluates LLMs' multimodal performance, called \textbf{MultiAPI}. Based on the HuggingFace dataset \cite{patil2023gorilla}, we extracted models with high-quality descriptions across 9 domains along with their instructions. These models were initially encapsulated as API functions using ChatGPT prompts, followed by meticulous human refinements to ensure executability and consistent argumentation across domains. This help create the \textbf{MultiAPI} benchmark dataset with 235 functional API calls and 2,038 instructions in this paper.

We subsequently conducted experiments on both API-based LLMs and open-sourced LLMs, exploring strategies that were previously proven effective in improving LLM prompting such as in-context learning \cite{brown2020language} and chain-of-thought \cite{wei2023chainofthought}. Our investigation spanned single-step API call (only 1 API is required to resolve the instruction) and sequential API chain (multiple APIs are required) settings, evaluating 4 intuitive aspects: 1) invocation assessment; 2) domain match; 3) function match; and 4) argument match. Results revealed that while models accurately make decisions to invoke API functions, they often suffer from selecting the right function and parameters from the correct domain. Furthermore, we surprisingly noticed that adding auxiliary context could harm the API call performance. Extensive error analyses were conducted to understand the potential cause of such errors, leading us to propose two simple yet effective solutions to mitigate these errors. The experimental results validate the effectiveness of our method. 

We summarize the contributions of this paper as follows:
\begin{itemize}
    \item We constructed a pioneering large-scale multimodal instruction-function benchmark dataset, \textbf{MultiAPI}, with 235 executable API functions and 2,038 prompts. This data underwent rigorous human refinement to ensure its robustness and relevance in the context of LLM evaluations.

    \item Our experimental framework comprehensively assesses both API-based and open-sourced LLMs, revealing their strengths in API call decisions but highlighting challenges in domain and function selection, as well as argument generation.

    \item A thorough error analysis leads us to mitigate these errors and set a new direction for future LLM research within the multimodal context.
\end{itemize}

\section{Related Work}
\subsection{Evaluation of Large Language Models}
Performance evaluation of LLMs has become a particularly prominent field postdate of the introduction of ChatGPT, providing valuable insights for enhancing future model iterations and assisting the industry in developing more resilient applications. Extensive research has been undertaken to assess the competencies of LLMs\cite{yin2023survey, yang2023dawn, laskar-etal-2023-systematic, zhang2023language}. These works demonstrated LLMs expressed near-human performance on open-domain tasks such as mathematics, coding, law, and psychology. However, their proficiency with tool use has not been thoroughly explored.

\citet{li2023apibank} introduced a benchmark for assessing LLMs' tool-use proficiency through a set of APIs. This benchmark, however, the amount of APIs is constrained by its reliance on human implementation and primarily evaluates LLMs on general tasks like setting alarms or scheduling meetings.

In contrast, our study pivots to evaluate LLMs' ability to handle multimodal tasks via the use of tool APIs. We have harnessed ChatGPT's code generation capabilities based on the provided code template, followed by meticulous human refinement, to construct \textbf{MultiAPI}, a high-quality and large-scale multimodal API dataset. This novel dataset enables us to delve into the multimodal task performance of LLMs, marking a significant advancement in the field.

\begin{figure*}[!ht]
    \centering
    \includegraphics[width=1.03\textwidth]{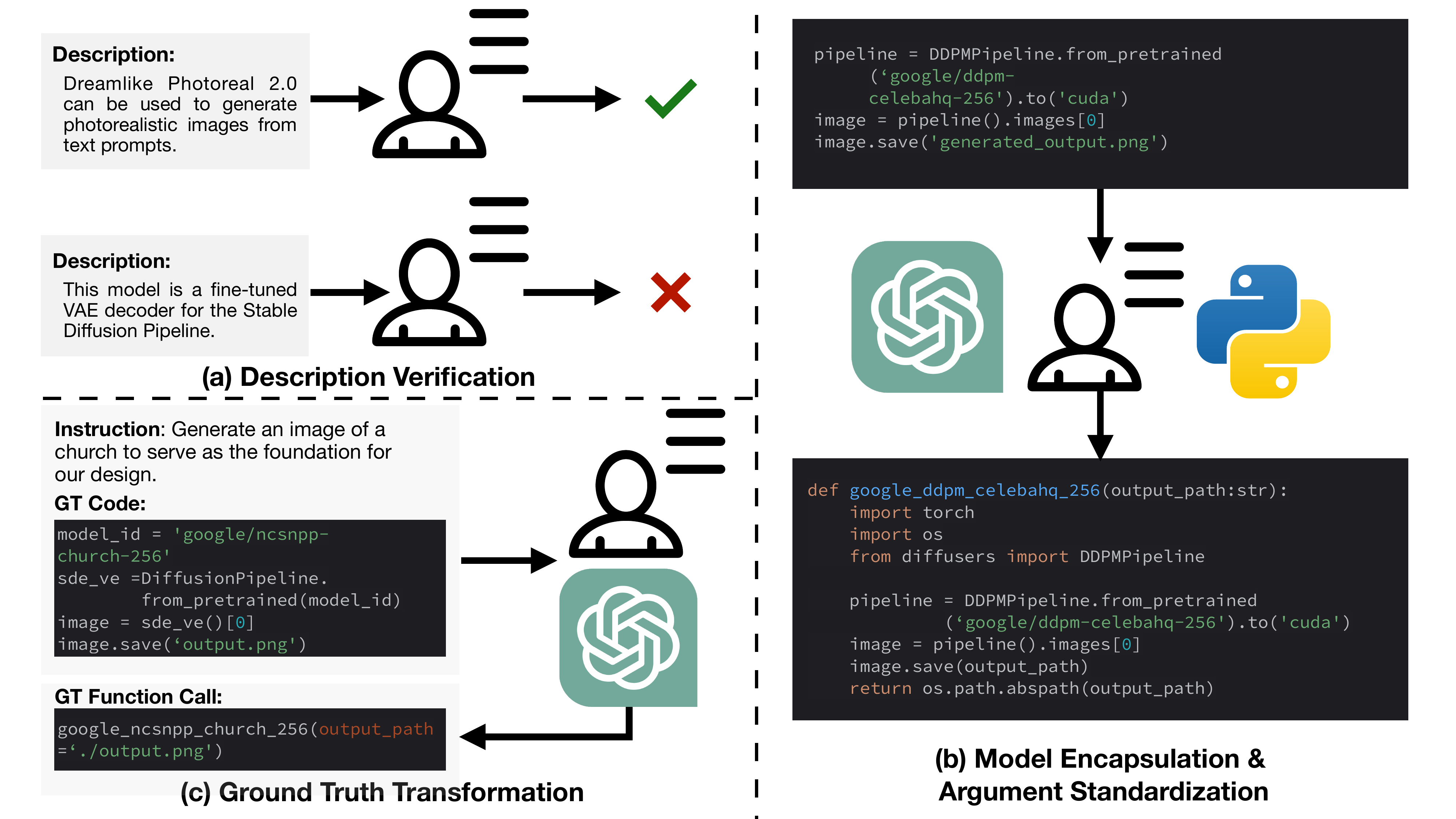}
    \caption{Workflow for adapting the HuggingFace dataset for MultiAPI collaboration with GPT model:  (a) the Description Verification process where model descriptions are assessed for precision and detail. (b) the Model Encapsulation and Argument Standardization procedure, transitioning from an 'example code' format to an argument-standardized Python function and ensuring the function is executable. (c) the Ground Truth Transformation, showing the conversion of instruction-code pairs into instruction-function pairs.}
    \label{fig:data_collection}
\end{figure*}

\subsection{Large Language Model Augmentation}
Although large language models recently demonstrated superior zero-shot language understanding \cite{openai2023gpt4, touvron2023llama, zhang2023extractive} capability, the task scope they could handle is highly tethered with their pretraining data. To adapt LLMs to diverse inputs and tasks, recent studies have primarily followed two avenues. The first involves joint fine-tuning of LLMs with pertinent neural network components. In this approach, the hidden representations of novel modalities are aligned with the LLM's latent space\cite{awais2023foundational, wu2023nextgpt, yin2023survey, patil2023gorilla, lyu2023macaw}. The second avenue integrates tools such as API functions as external modules\cite{schick2023toolformer, zhang2023graph, song2023restgpt}. The strategy offers enhanced flexibility, allowing API functions to be seamlessly incorporated into textual contexts, irrespective of whether the LLM is API-centric or open-sourced.

Several studies have examined combining large language models with external resources. \citet{shen2023hugginggpt} notably linked ChatGPT with HuggingFace, enhancing its decision-making range. However, this integration struggled with producing precise code due to inconsistencies in the ground truth code and insufficient documentation.  In our study, we mitigated these limitations by utilizing human annotators to integrate each HuggingFace model as a function call. We also standardized function arguments within the same domain, simplifying the evaluation process and reducing the complexity of model interactions during assessments.


\begin{table*}[!ht]
    \centering
    \small
    \begin{tabularx}{0.80\textwidth}{llc}
    \toprule
    \textbf{Domains} & \textbf{Required Arguments} & \textbf{\# Functions}\\
    \midrule
    Text to Image & \texttt{(prompt: str, output\_path:str)} & 11\\
    Depth Estimation & \texttt{(image\_path:str, output\_path:str)} & 10\\
    Object Detection & \texttt{(image\_path: str)} & 30\\
    Video Classification & \texttt{(video\_path:str)} & 23\\
    Image Classification & \texttt{(image\_path: str)} & 48\\
    Image to Text & \texttt{(image\_path: str)} & 28 \\
    Image Generation & \texttt{(output\_path:str)} & 33\\
    Image Segmentation & \texttt{(image\_path: str, prompt:str)} & 29\\
    Image to Image & \texttt{(control\_image\_path:str, output\_image\_path:str)}& 23 \\
    \bottomrule
    \end{tabularx}
    \caption{The domains of MultiAPI and their required arguments. \# Functions represents the number of functions that each domain contains.}
    \label{tab:domains}
\end{table*}

\section{MultiAPI Benchmark Dataset}
\subsection{Data Collection}

In this section, we detail the process of constructing \textbf{MultiAPI} leveraging the HuggingFace instruction-code dataset introduced by \citet{patil2023gorilla}. The original dataset consists of a model definition file including model descriptions along with its corresponding example code template; and an instruction-code pair file linking models to self-generated instructions\cite{wang2023selfinstruct}.

We first filtered out all the models that could potentially assist multimodal tasks from 9 unique domains, as shown in Table \ref{tab:domains}, and their corresponding instruction-code pairs. The subsequent data processing comprises four steps: 1) Description Verification, 2) Model Encapsulation, 3) Argument Standardization, and 4) Ground Truth Transformation. The primary procedures are illustrated in Figure \ref{fig:data_collection}. It's noteworthy that the first three steps are applied to the model definition and the last is applied to the instruction-code pair.

\paragraph{Description Verification:}While most models come equipped with a description field that provides the basic information, the quality of these descriptions varies widely, largely depending on community contributors. Previous studies verified that a precise and detailed model description plays a critical role in aiding the model to identify the appropriate tool based on user specifications \cite{hsieh2023tool}. Such specificity could also bolster the accuracy and reliability of evaluation outcomes. To this end, we engaged two human annotators with expertise in NLP to manually review all descriptions. They were tasked with removing the model whose descriptions only offered a broad overview, lacking a delineated use case, as depicted in lower Figure \ref{fig:data_collection} (a).

\paragraph{Model Encapsulation:} The primary utility of the original dataset was to facilitate the training or finetuning of LLMs to autonomously generate the API call code, contingent on retrieval results. Consequently, models were invoked using the \texttt{example\_code} field present in the dataset, as illustrated in the upper section of Figure \ref{fig:data_collection}(b). To adapt the existing models from HuggingFace to the API function-calling framework, we prompt \textit{gpt-3.5-turbo} to transform the example code template into an API function and subsequently extract the potential arguments. In addition, we identify and include the import statements inside the function to ensure the function is independently executable. 


\paragraph{Argument Standardization:}Upon encapsulating the functions, we observe that while \textit{gpt-3.5-turbo} adeptly transformed essential codes into function form, it exhibited challenges in accurately extracting function arguments. Further analysis suggests that the variation in argument names and the number of arguments pose a significant challenge \cite{yin2023woodpecker}, potentially introducing the risk of hallucination, ambiguity and complicating the parsing process during argument evaluations. To address the aforementioned discrepancies, we introduce an argument standardization process. Consider a function set $F_d$ within a given domain $d$. We define a standardized argument set $A_d$ by manually reviewing all functions within $d$ to determine the commonly recurring arguments intrinsic to the domain's functionality. As a result, for any functions within $d$, we require:
\begin{equation}
\forall f_1, f_2 \in F_d, \quad \text{args}(f_1) = \text{args}(f_2) = A_d
\end{equation}
For instance, within the \textit{Text to Image} domain, functions generate images in response to user prompts. Consequently, the indispensable arguments for this domain are {\texttt{prompt}} and {\texttt{output\_path}}. The detailed mappings between domains and required arguments are listed in Table \ref{tab:domains}.

Using this collated reference table, human experts are introduced to refine the generated functions ensuring: 1) Incorporation of the minimum required arguments, named in line with the reference table. 2) Listing other arguments as default arguments with default values. 3) Ensuring the function remains executable within Python environments.

\paragraph{Ground Truth Transformation:}As shown in the upper segment of Figure \ref{fig:data_collection}(c), instruction-code pairs represent specific instructions with their corresponding code blocks. To maintain consistency with our previous steps, we use a similar human-supervised approach with \textit{gpt-3.5-turbo} to transform these pairs into instruction-function pairs. The results are depicted in the bottom code block of Figure \ref{fig:data_collection}(c). This ensures a streamlined and consistent framework for both model definitions and their corresponding instructions.

\subsection{Evaluation Metrics}
\label{sec: eval_metrics}
The outputs of multimodal tasks are contingent on varying input modalities, leading to unpredictable results even with identical inputs \cite{rombach2022high, saharia2022photorealistic}, which makes direct evaluation on output unreliable. Moreover, crafting robust evaluation metrics for each individual domain poses significant challenges for future versatility. 
However, benefiting from diligent data collection steps, we bypass these issues by assessing the LLM's tool usage ability based on the function calls selected. In function-calling context, user's requirement would be fulfilled if the model correctly selects the appropriate function and fills in the accurate arguments. This approach streamlines the evaluation into a universal domain-agnostic text-matching task with some necessary adaptions.  

Inspired by \citet{li2023apibank}, we design a step-wise, four-level evaluation framework for a comprehensive assessment of LLMs' tool usage in multimodal tasks. This framework includes:
\begin{enumerate}
\item \textbf{Invocation Assessment:} Tests if LLMs can discern when a user instruction necessitates an auxiliary function.
\item \textbf{Domain Match:} Evaluates the LLMs' ability to match the function's domain to the ground truth by leveraging domain annotations in our dataset.
\item \textbf{Function Match:} Conducts a detailed assessment to confirm whether the LLM correctly identifies the specific tool within the matched domain via their descriptions.
\item \textbf{Argument Match:} Verifies the LLM's proficiency in translating user instructions into precise arguments for successful function invocation. The distinction in evaluating multimodal task functions lies in the API arguments. We classify arguments defined in Table \ref{tab:domains} into two distinct categories: exact-match arguments and concept-match arguments. Exact-match arguments, such as file paths, demand precise, verbatim replication. Any deviation in these arguments can impede the successful invocation of the function. On the other hand, concept-match arguments, like generative prompts, offer more flexibility in wording, though they must maintain fidelity in conveying the intended meaning. Inaccuracies in generating concept-match arguments, while not hindering the function invocation, can lead to outputs that diverge from the expected results. 

In our experiments, exact-match arguments undergo text matching for exact path alignment, while concept-match prompts are semantically evaluated using ROUGE F-scores \cite{lin2004rouge} and cosine similarity \cite{7577578} for both statistical and vectorized analysis. 

\end{enumerate}





\section{Experiments}
\label{sec:4exps}
In this section, we conduct extensive experiments on our proposed MutilAPI benchmark to assess the capabilities of LLMs in handling multimodal tasks through tool integration. Our evaluation spans both API-based models and open-source models. For each model, we implement a variety of prompt configurations, aiming to identify the most effective prompt settings specifically tailored for multimodal tasks.

\subsection{Task Formulation}
Given a multimodal task instruction $i$, the model's objective is to generate an API function $f$ from a set of available functions $F$ and its corresponding set of arguments $A_f$. Formally, for $f \in F$ the generation process can be represented as:

\begin{equation}
p(f, A|i, F) = p(f|i, F) \times p(A| f, i)
\end{equation}


\subsection{Models and Prompt Configurations}
Current LLMs can be categorized into API-based models and open-sourced models. Our evaluation performs on both categories. For API-based models, we use \textit{gpt-3.5-turbo-0613} as the candidate. For open-sourced models, we leverage Llama2-13B \cite{touvron2023llama} provided by HuggingFace\footnote{https://huggingface.co/docs/transformers/main/model\_doc/llama2}. Furthermore, previous research proved prompt configurations can significantly affect the performance of LLMs \cite{zhang2023summit, wei2022chain}. To investigate whether these configurations remain effective on our task. We implemented the following prompt configurations in our experiments:

\paragraph{In-context Learning:} Previous research demonstrated the few-shot performance of language models can be significantly boosted by providing exemplar input-ground truth pairs \cite{brown2020language}. In our in-context setting, we provide 2 instruction-function call pairs to assist the model in reasoning the predictions.

\paragraph{Chain-of-Thought:} Chain-of-Thought \cite{wei2023chainofthought} adapts the concept of divide-and-conquer. It allows LLMs to address problems in a step-by-step fashion, by deconstructing the primary task into smaller, manageable queries. This approach not only simplifies the task but also bolsters the reasoning capabilities of the models. We apply this framework by breaking down the task into 4 questions aligned with our evaluation metrics introduced in \ref{sec: eval_metrics}. 

\paragraph{Function Calling:} Recently introduced by OpenAI\footnote{https://platform.openai.com/docs/guides/function-calling}, Function Calling is a feature tailored for GPT models. The models are finetuned on a specialized function-call dataset. The intent is to enable the models to better recognize scenarios necessitating function calls, thereby facilitating the generation of more structured outputs.

\subsection{Context Token Limitation}
Given the constraint of a maximum context window of 4,096 tokens for those LLMs used in our experiments, we face a limitation in the number of functions that can be included within this token budget. Our calculations suggest that approximately 25 functions can be accommodated. To effectively manage this constraint, we initially shuffle the entire dataset. Subsequently, we divide it into 10 segments, each containing 25 functions, except for the final segment which may vary in size due to the distribution of the remaining functions. For each experiment configuration, we conduct separate trials on each of these 10 splits. The overall experimental results are then derived by calculating the average across these 10 segments.

\subsection{Function Invocation}
\label{sec:func_inv}

In this section, we will focus on the function invocation aspect of LLMs to evaluate their ability to understand user instructions and locate the proper tool function. The results are demonstrated in Table \ref{tab:main_exp}.

\paragraph{LLMs face challenges in multimodal domain selection:} By observing across columns, we could conclude both GPT-3.5 and Llama models exhibit commendable accuracy in determining the necessity of function invocation based on user instructions. However, a significant drop in performance occurs when it comes to identifying the specific domain of multimodal tasks and selecting the precise function to effectively address these tasks. This finding implies that, while LLMs possess robust common-sense knowledge, they still struggle with accurately comprehending the nuances and definitions unique to each domain of multimodal tasks. 


\begin{table}[t!]
\small
\centering
\scalebox{0.98}{
\begin{tabular}{l|ccc}
\toprule
\textbf{Model} & \begin{tabular}{@{}c@{}}\textbf{Invoke}  \\ \textbf{Accuracy}\end{tabular} & \begin{tabular}{@{}c@{}}\textbf{Domain}  \\ \textbf{Accuracy}\end{tabular} & \begin{tabular}{@{}c@{}}\textbf{Function}  \\ \textbf{Accuracy}\end{tabular} \\
\midrule
GPT-3.5 & 99.82 & \textbf{71.78} & \textbf{52.94} \\
GPT-3.5-cot & \textbf{99.95} & 71.43 & 51.73 \\
GPT-3.5-ict & 99.47 & 68.07 & 48.35 \\
GPT-3.5-ict-cot & 98.77 & 64.00 & 48.16 \\
\midrule
GPT-3.5-fc & \textbf{99.11} & \textbf{75.52} & \textbf{55.53} \\
GPT-3.5-fc-cot & 94.13 & 70.00 & 50.12 \\
GPT-3.5-fc-ict & 95.02 & 67.72 & 49.59 \\
GPT-3.5-fc-ict-cot & 98.41 & 69.91 & 51.62 \\
\midrule
Llama & 85.87 & \textbf{14.75} & \textbf{9.94} \\
Llama-cot & 79.88 & 12.76 & 6.37 \\
Llama-ict & 83.59 & 10.70 & 5.72 \\
Llama-ict-cot & \textbf{86.30} & 10.56 & 5.00 \\ 
\bottomrule
\end{tabular}
}
\caption{Experimental results for function selection across different LLM configurations, where '-cot' denotes the use of Chain-of-Thought prompting, '-incontext' signifies incontext learning, and '-fc' indicates that the function calling feature is enabled.}
\label{tab:main_exp}
\end{table}

\begin{table}[t!]
\small
\centering
\scalebox{0.83}{
\begin{tabular}{l|c|ccccc}
\toprule
\textbf{Model} & \begin{tabular}{@{}c@{}}\textbf{Argument}  \\ \textbf{Accuracy}\end{tabular} & \textbf{R1} & \textbf{R2} & \textbf{RL} & \textbf{Sim} \\
\midrule
GPT-3.5 & \textbf{42.68} & 25.05 & 17.94 & 24.64 & 46.61 \\
GPT-3.5-ict & 36.37 & 30.37 & 21.32 & 29.68 & 50.82 \\
GPT-3.5-cot & 41.12 & 24.84 & 17.81 & 24.31 & 46.39 \\
GPT-3.5-ict-cot & 25.79 & \textbf{32.45} & \textbf{22.78} & \textbf{31.95} & \textbf{53.97} \\
\midrule
GPT-3.5-fc & \textbf{43.40 }& 24.17 & 15.42 & 23.39 & 44.64 \\
GPT-3.5-fc-ict & 32.26 & \textbf{24.67} & \textbf{16.63} & \textbf{24.10} & 44.05 \\
GPT-3.5-fc-cot & 38.26 & 24.53 & 15.45 & 23.85 & \textbf{45.50} \\
GPT-3.5-fc-ict-cot & 18.91 & 23.65 & 15.09 & 22.86 & 45.14 \\
\bottomrule
\end{tabular}
}
\caption{Comparative evaluation of GPT-3.5 model configurations in argument generation. The first section shows the match accuracy of exact-match arguments while the second demonstrate the evaluation metrics of concept-match parameters. R1/2/L represents ROUGE-1/2/L scores respectively, and Sim represents cosine similarity.}
\label{tab:parameter_comparison}
\end{table}

\paragraph{Function Calling enhancement performance varied by prompt configuration: } Upon comparing the results in the first and second blocks of Table \ref{tab:main_exp}, it is evident that enabling Function Calling significantly enhances performance in the GPT-3.5 and GPT-3.5-ict-cot configurations, while it appears to slightly impede performance in settings where only a single prompt configuration is employed. This observation could potentially be attributed to the complex interplay between the Function Calling mechanism and the prompt configurations. Such findings underscore the importance of carefully considering the compatibility of various features and configurations when augmenting LLMs for specific tasks.

\paragraph{In-context learning impairs multimodal function invocation: } Our analysis of the effectiveness of prompt configurations, conducted through a cross-row examination within each block, revealed consistent patterns across both GPT-3.5 and Llama models. A prominent observation is that the incorporation of contextual elements tends to negatively impact performance, a trend that is especially pronounced with the introduction of in-context learning. This significant impairment in performance is contrary to the widespread belief that providing reference context generally improves model performance across a variety of tasks. Such a result suggests that in multimodal function invocation scenarios, the addition of contextual information might inadvertently introduce complexity or irrelevant data, thus diminishing the model’s efficiency. This counterintuitive finding points to a need for deeper inquiry into how and why the incorporation of contextual elements in LLMs affects their function invocation capabilities, challenging existing assumptions and opening new avenues for research in the field.

\subsection{Argument Generation}


The capabilities of LLMs in generating arguments for multimodal tasks are detailed in Table \ref{tab:parameter_comparison}. It's noteworthy that Llama was excluded from this analysis due to its inferior performance in function locating. The results indicate a significant challenge for GPT models in accurately generating both exact-match and concept-match arguments based on user instructions. The success rate for matching exact-match arguments falls below 50\%, and the semantic similarity of the generated concept-match arguments is similarly subpar. This suggests that argument generation set a more critical bottleneck hindering LLMs' ability to effectively invoke multimodal functions, compared to the function invocation ability in the previous sections.

Additionally, a distinct observation from the data is that, while the exact-match argument accuracy aligns with previous insights, the inclusion of additional context appears to positively impact the generation of concept-match arguments. This highlights a nuanced aspect of LLM performance, where contextual information plays a more beneficial role in generating arguments that rely on semantic rather than verbatim accuracy. This finding suggests potential areas for optimization in LLMs, particularly in enhancing their ability to handle concept-match arguments in the context of multimodal task execution while reducing the performance impairment on the exact-match arguments.

\begin{table}[t]
\centering
\small
\scalebox{1.05}{
\begin{tabular}{l|cccc}
\toprule
& \multicolumn{2}{c}{\textbf{GPT-3.5}} & \multicolumn{2}{c}{\textbf{GPT-3.5-fc}} \\
\cmidrule(lr){2-3} \cmidrule(lr){4-5}
\textbf{Metric} & \textbf{Func 1} & \textbf{Func 2} & \textbf{Func 1} & \textbf{Func 2} \\
\midrule
\textbf{Inv Acc} & 99.76 & 99.94 & 99.83 & 60.00 \\
\textbf{Dm Acc} & 76.67 & 36.67 & 66.67 & 40.00 \\
\textbf{Func Acc} & 53.33 & 30.00 & 46.67 & 40.00 \\
\midrule
\textbf{Arg Acc} & 86.36 & 31.25 & 89.47 & 61.54 \\
\textbf{R1} & 63.17 & 67.25 & 69.68 & 68.57 \\
\textbf{R2} & 45.58 & 64.30 & 54.60 & 54.00 \\
\textbf{RL} & 61.15 & 67.25 & 69.67 & 67.69 \\
\textbf{Sim} & 83.28 & 75.69 & 86.68 & 70.45 \\
\bottomrule
\end{tabular}
}
\caption{Sequential API invocation result on \textbf{MultiAPI-SEQ}. The metrics evaluated include Invocation Accuracy (Inv Acc), Domain Accuracy (Dm Acc), Function Accuracy (Func Acc), Argument Accuracy (Arg Acc), ROUGE-1/2/L (R1/2/L), Similarity (Sim).}
\label{tab:seq_call}
\end{table}

\subsection{Sequential API Invocation}
In real-world applications, user instructions frequently necessitate the invocation of multiple APIs for resolution. Particularly in our multimodal scenario, this requires LLMs to possess a thorough understanding of each modality and associated tasks, as well as the interaction between modalities. Analyzing models' capabilities in sequential API invocation is more representative of real-life applications and offers valuable insights for application development. To address this need, we introduce \textbf{MultiAPI-SEQ}, a dataset specifically designed for assessing sequential function invocation. This dataset has been carefully curated by human experts who have manually crafted 30 distinct instructions. Each of these instructions necessitates the sequential invocation of two functions from the \textbf{MultiAPI} dataset. By limiting each instruction to require just two functions, we aim to simplify the analysis process while still effectively evaluating the models' ability to handle multi-step task execution. 

As shown in Table \ref{tab:seq_call}, the analysis of GPT-3.5 and GPT-3.5-fc models demonstrate significant inconsistency in maintaining performance across sequential tasks. Both models exhibit high invocation accuracy initially, yet GPT-3.5-fc's accuracy notably diminishes during the second task. This indicates that while fine-tuning may enhance single-function call performance, it could adversely affect task planning in sequential API call tasks. This trend toward relying on built-in parametric knowledge instead of external tools raises concerns about the potential for hallucination. Additionally, both models show a reduction in domain and function accuracy, with GPT-3.5-fc's argument accuracy notably less affected, implying a relatively stable understanding of argument relevance. The linguistic similarity metrics across functionalities indicate that GPT-3.5 demonstrates more consistent performance, hinting at its robustness in generating contextually appropriate responses throughout the task sequence.

\section{Analysis}
\label{sec:5analysis}
To investigate the potential causes of LLMs' underperformance, we performed a detailed error analysis at the domain and function levels for GPT-3.5-fc.
\begin{figure}[t!]
    \centering
    \includegraphics[width=0.45\textwidth]{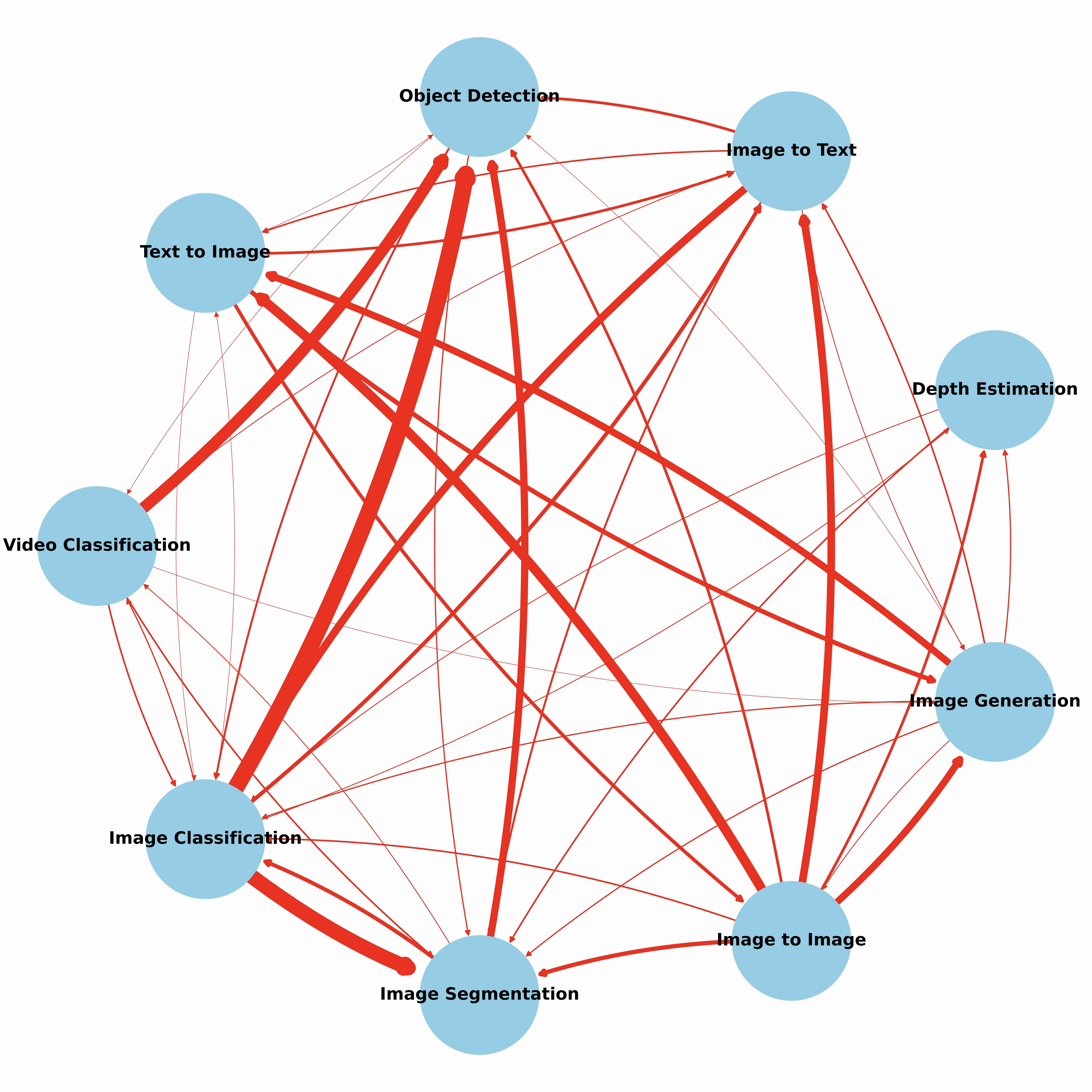}
    \caption{Domain misclassification network. Nodes in this graph represent distinct domains, with directed arrows illustrating instances where the model incorrectly applies a function from domain $b$ intended for an instruction in domain $a$. The thickness of the arrows indicates the frequency of these errors, with thicker lines showing more common misclassifications.}
    \label{fig:domain_mis}
\end{figure}

\begin{figure}[t!]
    \centering
    \includegraphics[width=0.5\textwidth]{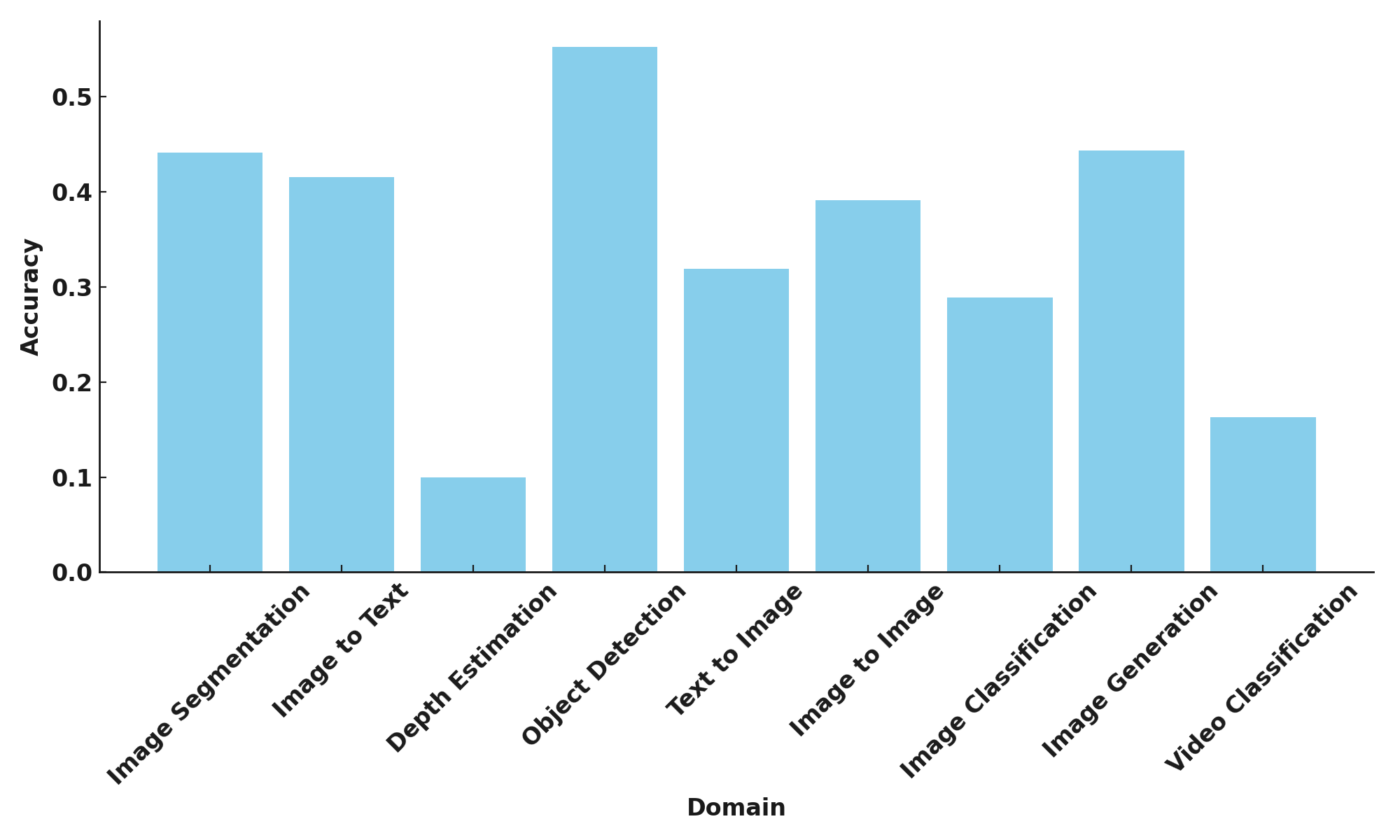}
    \caption{Function accuracy distribution for each domain.}
    \label{fig:func_mis}
\end{figure}

\subsection{Domain Mismatch}

Section \ref{sec:func_inv} suggests LLMs struggle to differentiate multimodal task domains. We analyze model errors to identify these shortcomings. We summarize the result as a misclassification network indicating LLM's domain confusion in Figure \ref{fig:domain_mis}.

For visual analysis APIs, the model demonstrates an inclination to misinterpret classification and segmentation tasks as object detection. Besides, it also frequently fails the identification between image classification and image segmentation. This pattern indicates a fundamental challenge in the LLM's ability to identify domains based on user instruction, particularly in discerning whether the analysis should encompass the entire image or focus on the specific content within the image. The asymmetries in bidirectional error between these nodes further suggest that LLM bias towards local rather than global image analysis.

Additionally, with image generation APIs, the model often struggles to determine whether a task is conditional or unconditional, commonly misidentifying text-to-image and image-to-image tasks as unconditional image generation. It also faces challenges in recognizing the input modality for conditioned generation tasks, as evidenced by errors between image-to-image and text-to-image tasks. Those two observations may suggest that LLMs lack an understanding of different modalities, possibly because they are predominantly trained on textual data.

\subsection{Function Mismatch}
To assess the LLMs' function selection accuracy, we randomly sampled 10 functions and corresponding instructions from each domain and prompted the model to choose the most appropriate function within that domain. As shown in Figure \ref{fig:domain_mis}, the histogram reflecting function accuracy across domains, demonstrates the uneven function selection proficiency of LLMs in handling different multimodal tasks. Domains with more straightforward, visually dense tasks like image-to-image and object detection demonstrate relatively high accuracy, indicating that models perform better with tasks requiring less complex language-to-function mapping. In contrast, the low accuracy in depth estimation' and video classification points to the models' limitations in understanding and translating more abstract or dynamic task requirements into accurate function calls.

\section{Improvement Framework}

\begin{table}[t]
\centering
\small
\scalebox{1.05}{
\begin{tabular}{l|cc}
\toprule
\textbf{Metric} & \textbf{GPT-3.5} & \textbf{GPT-3.5-dp-ac} \\
\midrule
\textbf{Inv Acc} & 99.82 & \textbf{99.87} \\
\textbf{Dm Acc} & 71.78 & \textbf{76.31} \\
\textbf{Func Acc} & 51.73 & \textbf{59.47} \\
\midrule
\textbf{Arg Acc} & 42.68 & \textbf{48.82} \\
\textbf{R1} & 25.05 & \textbf{27.76} \\
\textbf{R2} & 17.94 & \textbf{18.45} \\
\textbf{RL} & 24.64 & \textbf{26.33} \\
\textbf{Sim} & 46.61 & \textbf{56.82} \\
\bottomrule
\end{tabular}
}
\caption{The experiment result of adding detailed domain description prompting (-dp) and argument correction (-ac).}
\label{tab:imporve}
\end{table}
Our analysis in Sections \ref{sec:4exps} and \ref{sec:5analysis} reveals that LLMs primarily struggle with distinguishing domain differences and modalities, with argument generation as a significant bottleneck. To mitigate these challenges, we propose two intuitive yet effective solutions: domain description prompting and argument revision.

Domain description prompting involves adding a sentence to the model's system prompt to clearly define each domain. In addition, in visual analysis tasks, we specify whether the domain conducts global or local image analysis. 

Building on research showing LLMs' effectiveness in evaluation and revision tasks \cite{liu2023gpteval, zhang2023summit}, we employ a secondary LLM as an argument editor. This LLM checks and revises argument predictions to ensure they align with user instructions, reducing task complexity and the context length for the primary LLM.

To avoid the noise arising from complex interactions between function calling feature and input context, we conducted our experiments using the GPT-3.5 model. Table \ref{tab:imporve} illustrates that our approach enhanced performance across all evaluation metrics. Notably, there was a significant improvement in domain accuracy, argument exact matching, and semantic evaluation. This significant improvement not only affirms the effectiveness of our approach but also strongly validates the accuracy of our analysis. Furthermore, we observed a notable enhancement in function accuracy, attributed to the incorporation of domain descriptions.

\section{Conclusion}
In this paper, we presented a comprehensive study on the application of Large Language Models to multimodal tasks with external API functions, using the newly introduced \textbf{MultiAPI} dataset. Our findings highlight the capabilities and limitations of LLMs in function calling. We revealed a significant discrepancy between the models' ability to recognize the need for function calls and their accuracy in selecting appropriate domains, functions, and arguments. This insight led us to propose a novel approach focusing on domain description prompting and argument revision, which demonstrated improved performance in addressing these challenges. Our work contributes to the field by introducing the first large-scale multimodal instruction-function benchmark dataset and providing a detailed analysis of LLMs in multimodal task execution. We hope our dataset and findings could assist the development of tool-augmented LLMs and more sophisticated models for complex real-world applications. 

\newpage




\bibliography{anthology,custom}

\begin{thebibliography}{37}
\expandafter\ifx\csname natexlab\endcsname\relax\def\natexlab#1{#1}\fi

\bibitem[{Awais et~al.(2023)Awais, Naseer, Khan, Anwer, Cholakkal, Shah, Yang, and Khan}]{awais2023foundational}
Muhammad Awais, Muzammal Naseer, Salman Khan, Rao~Muhammad Anwer, Hisham Cholakkal, Mubarak Shah, Ming-Hsuan Yang, and Fahad~Shahbaz Khan. 2023.
\newblock Foundational models defining a new era in vision: A survey and outlook.
\newblock \emph{arXiv preprint arXiv:2307.13721}.

\bibitem[{Brown et~al.(2020)Brown, Mann, Ryder, Subbiah, Kaplan, Dhariwal, Neelakantan, Shyam, Sastry, Askell et~al.}]{brown2020language}
Tom Brown, Benjamin Mann, Nick Ryder, Melanie Subbiah, Jared~D Kaplan, Prafulla Dhariwal, Arvind Neelakantan, Pranav Shyam, Girish Sastry, Amanda Askell, et~al. 2020.
\newblock Language models are few-shot learners.
\newblock \emph{Advances in neural information processing systems}, 33:1877--1901.

\bibitem[{Bubeck et~al.(2023)Bubeck, Chandrasekaran, Eldan, Gehrke, Horvitz, Kamar, Lee, Lee, Li, Lundberg, Nori, Palangi, Ribeiro, and Zhang}]{bubeck2023sparks}
Sébastien Bubeck, Varun Chandrasekaran, Ronen Eldan, Johannes Gehrke, Eric Horvitz, Ece Kamar, Peter Lee, Yin~Tat Lee, Yuanzhi Li, Scott Lundberg, Harsha Nori, Hamid Palangi, Marco~Tulio Ribeiro, and Yi~Zhang. 2023.
\newblock \href {http://arxiv.org/abs/2303.12712} {Sparks of artificial general intelligence: Early experiments with gpt-4}.

\bibitem[{Hsieh et~al.(2023)Hsieh, Chen, Li, Fujii, Ratner, Lee, Krishna, and Pfister}]{hsieh2023tool}
Cheng-Yu Hsieh, Si-An Chen, Chun-Liang Li, Yasuhisa Fujii, Alexander Ratner, Chen-Yu Lee, Ranjay Krishna, and Tomas Pfister. 2023.
\newblock Tool documentation enables zero-shot tool-usage with large language models.
\newblock \emph{arXiv preprint arXiv:2308.00675}.

\bibitem[{Lahitani et~al.(2016)Lahitani, Permanasari, and Setiawan}]{7577578}
Alfirna~Rizqi Lahitani, Adhistya~Erna Permanasari, and Noor~Akhmad Setiawan. 2016.
\newblock \href {https://doi.org/10.1109/CITSM.2016.7577578} {Cosine similarity to determine similarity measure: Study case in online essay assessment}.
\newblock In \emph{2016 4th International Conference on Cyber and IT Service Management}, pages 1--6.

\bibitem[{Laskar et~al.(2023)Laskar, Bari, Rahman, Bhuiyan, Joty, and Huang}]{laskar-etal-2023-systematic}
Md~Tahmid~Rahman Laskar, M~Saiful Bari, Mizanur Rahman, Md~Amran~Hossen Bhuiyan, Shafiq Joty, and Jimmy Huang. 2023.
\newblock \href {https://doi.org/10.18653/v1/2023.findings-acl.29} {A systematic study and comprehensive evaluation of {C}hat{GPT} on benchmark datasets}.
\newblock In \emph{Findings of the Association for Computational Linguistics: ACL 2023}, pages 431--469, Toronto, Canada. Association for Computational Linguistics.

\bibitem[{Li et~al.(2023{\natexlab{a}})Li, Gan, Yang, Yang, Li, Wang, and Gao}]{li2023multimodal}
Chunyuan Li, Zhe Gan, Zhengyuan Yang, Jianwei Yang, Linjie Li, Lijuan Wang, and Jianfeng Gao. 2023{\natexlab{a}}.
\newblock \href {http://arxiv.org/abs/2309.10020} {Multimodal foundation models: From specialists to general-purpose assistants}.

\bibitem[{Li et~al.(2023{\natexlab{b}})Li, Zhao, Yu, Song, Li, Yu, Li, Huang, and Li}]{li2023apibank}
Minghao Li, Yingxiu Zhao, Bowen Yu, Feifan Song, Hangyu Li, Haiyang Yu, Zhoujun Li, Fei Huang, and Yongbin Li. 2023{\natexlab{b}}.
\newblock \href {http://arxiv.org/abs/2304.08244} {Api-bank: A comprehensive benchmark for tool-augmented llms}.

\bibitem[{Li et~al.(2023{\natexlab{c}})Li, Yang, and Wang}]{li2023reinforcement}
Zihao Li, Zhuoran Yang, and Mengdi Wang. 2023{\natexlab{c}}.
\newblock \href {http://arxiv.org/abs/2305.18438} {Reinforcement learning with human feedback: Learning dynamic choices via pessimism}.

\bibitem[{Lin(2004)}]{lin2004rouge}
Chin-Yew Lin. 2004.
\newblock Rouge: A package for automatic evaluation of summaries.
\newblock In \emph{Text summarization branches out}, pages 74--81.

\bibitem[{Liu et~al.(2023)Liu, Iter, Xu, Wang, Xu, and Zhu}]{liu2023gpteval}
Yang Liu, Dan Iter, Yichong Xu, Shuohang Wang, Ruochen Xu, and Chenguang Zhu. 2023.
\newblock Gpteval: Nlg evaluation using gpt-4 with better human alignment.
\newblock \emph{arXiv preprint arXiv:2303.16634}.

\bibitem[{Lyu et~al.(2023)Lyu, Wu, Wang, Huang, Liu, Du, Shi, and Tu}]{lyu2023macaw}
Chenyang Lyu, Minghao Wu, Longyue Wang, Xinting Huang, Bingshuai Liu, Zefeng Du, Shuming Shi, and Zhaopeng Tu. 2023.
\newblock Macaw-llm: Multi-modal language modeling with image, audio, video, and text integration.
\newblock \emph{arXiv preprint arXiv:2306.09093}.

\bibitem[{Maaz et~al.(2023)Maaz, Rasheed, Khan, and Khan}]{maaz2023videochatgpt}
Muhammad Maaz, Hanoona Rasheed, Salman Khan, and Fahad~Shahbaz Khan. 2023.
\newblock \href {http://arxiv.org/abs/2306.05424} {Video-chatgpt: Towards detailed video understanding via large vision and language models}.

\bibitem[{OpenAI(2023)}]{openai2023gpt4}
OpenAI. 2023.
\newblock \href {http://arxiv.org/abs/2303.08774} {Gpt-4 technical report}.

\bibitem[{Patil et~al.(2023)Patil, Zhang, Wang, and Gonzalez}]{patil2023gorilla}
Shishir~G. Patil, Tianjun Zhang, Xin Wang, and Joseph~E. Gonzalez. 2023.
\newblock \href {http://arxiv.org/abs/2305.15334} {Gorilla: Large language model connected with massive apis}.

\bibitem[{Qin et~al.(2023)Qin, Liang, Ye, Zhu, Yan, Lu, Lin, Cong, Tang, Qian et~al.}]{qin2023toolllm}
Yujia Qin, Shihao Liang, Yining Ye, Kunlun Zhu, Lan Yan, Yaxi Lu, Yankai Lin, Xin Cong, Xiangru Tang, Bill Qian, et~al. 2023.
\newblock Toolllm: Facilitating large language models to master 16000+ real-world apis.
\newblock \emph{arXiv preprint arXiv:2307.16789}.

\bibitem[{Qiu et~al.(2023)Qiu, Zhu, Han, Kumar, Mittal, Jin, Yang, Li, Wang, Li, Zhao, and Wang}]{qiu2023multisum}
Jielin Qiu, Jiacheng Zhu, William Han, Aditesh Kumar, Karthik Mittal, Claire Jin, Zhengyuan Yang, Linjie Li, Jianfeng Wang, Bo~Li, Ding Zhao, and Lijuan Wang. 2023.
\newblock \href {http://arxiv.org/abs/2306.04216} {Multisum: A dataset for multimodal summarization and thumbnail generation of videos}.

\bibitem[{Rombach et~al.(2022)Rombach, Blattmann, Lorenz, Esser, and Ommer}]{rombach2022high}
Robin Rombach, Andreas Blattmann, Dominik Lorenz, Patrick Esser, and Bj{\"o}rn Ommer. 2022.
\newblock High-resolution image synthesis with latent diffusion models.
\newblock In \emph{Proceedings of the IEEE/CVF Conference on Computer Vision and Pattern Recognition}, pages 10684--10695.

\bibitem[{Saharia et~al.(2022)Saharia, Chan, Saxena, Li, Whang, Denton, Ghasemipour, Gontijo~Lopes, Karagol~Ayan, Salimans et~al.}]{saharia2022photorealistic}
Chitwan Saharia, William Chan, Saurabh Saxena, Lala Li, Jay Whang, Emily~L Denton, Kamyar Ghasemipour, Raphael Gontijo~Lopes, Burcu Karagol~Ayan, Tim Salimans, et~al. 2022.
\newblock Photorealistic text-to-image diffusion models with deep language understanding.
\newblock \emph{Advances in Neural Information Processing Systems}, 35:36479--36494.

\bibitem[{Schick et~al.(2023)Schick, Dwivedi-Yu, Dess{\`\i}, Raileanu, Lomeli, Zettlemoyer, Cancedda, and Scialom}]{schick2023toolformer}
Timo Schick, Jane Dwivedi-Yu, Roberto Dess{\`\i}, Roberta Raileanu, Maria Lomeli, Luke Zettlemoyer, Nicola Cancedda, and Thomas Scialom. 2023.
\newblock Toolformer: Language models can teach themselves to use tools.
\newblock \emph{arXiv preprint arXiv:2302.04761}.

\bibitem[{Shen et~al.(2023)Shen, Song, Tan, Li, Lu, and Zhuang}]{shen2023hugginggpt}
Yongliang Shen, Kaitao Song, Xu~Tan, Dongsheng Li, Weiming Lu, and Yueting Zhuang. 2023.
\newblock \href {http://arxiv.org/abs/2303.17580} {Hugginggpt: Solving ai tasks with chatgpt and its friends in hugging face}.

\bibitem[{Song et~al.(2023)Song, Xiong, Zhu, Li, Wang, Tian, and Li}]{song2023restgpt}
Yifan Song, Weimin Xiong, Dawei Zhu, Cheng Li, Ke~Wang, Ye~Tian, and Sujian Li. 2023.
\newblock Restgpt: Connecting large language models with real-world applications via restful apis.
\newblock \emph{arXiv preprint arXiv:2306.06624}.

\bibitem[{Touvron et~al.(2023)Touvron, Martin, Stone, Albert, Almahairi, Babaei, Bashlykov, Batra, Bhargava, Bhosale, Bikel, Blecher, Ferrer, Chen, Cucurull, Esiobu, Fernandes, Fu, Fu, Fuller, Gao, Goswami, Goyal, Hartshorn, Hosseini, Hou, Inan, Kardas, Kerkez, Khabsa, Kloumann, Korenev, Koura, Lachaux, Lavril, Lee, Liskovich, Lu, Mao, Martinet, Mihaylov, Mishra, Molybog, Nie, Poulton, Reizenstein, Rungta, Saladi, Schelten, Silva, Smith, Subramanian, Tan, Tang, Taylor, Williams, Kuan, Xu, Yan, Zarov, Zhang, Fan, Kambadur, Narang, Rodriguez, Stojnic, Edunov, and Scialom}]{touvron2023llama}
Hugo Touvron, Louis Martin, Kevin Stone, Peter Albert, Amjad Almahairi, Yasmine Babaei, Nikolay Bashlykov, Soumya Batra, Prajjwal Bhargava, Shruti Bhosale, Dan Bikel, Lukas Blecher, Cristian~Canton Ferrer, Moya Chen, Guillem Cucurull, David Esiobu, Jude Fernandes, Jeremy Fu, Wenyin Fu, Brian Fuller, Cynthia Gao, Vedanuj Goswami, Naman Goyal, Anthony Hartshorn, Saghar Hosseini, Rui Hou, Hakan Inan, Marcin Kardas, Viktor Kerkez, Madian Khabsa, Isabel Kloumann, Artem Korenev, Punit~Singh Koura, Marie-Anne Lachaux, Thibaut Lavril, Jenya Lee, Diana Liskovich, Yinghai Lu, Yuning Mao, Xavier Martinet, Todor Mihaylov, Pushkar Mishra, Igor Molybog, Yixin Nie, Andrew Poulton, Jeremy Reizenstein, Rashi Rungta, Kalyan Saladi, Alan Schelten, Ruan Silva, Eric~Michael Smith, Ranjan Subramanian, Xiaoqing~Ellen Tan, Binh Tang, Ross Taylor, Adina Williams, Jian~Xiang Kuan, Puxin Xu, Zheng Yan, Iliyan Zarov, Yuchen Zhang, Angela Fan, Melanie Kambadur, Sharan Narang, Aurelien Rodriguez, Robert Stojnic, Sergey Edunov, and Thomas
  Scialom. 2023.
\newblock \href {http://arxiv.org/abs/2307.09288} {Llama 2: Open foundation and fine-tuned chat models}.

\bibitem[{Wang et~al.(2023)Wang, Kordi, Mishra, Liu, Smith, Khashabi, and Hajishirzi}]{wang2023selfinstruct}
Yizhong Wang, Yeganeh Kordi, Swaroop Mishra, Alisa Liu, Noah~A. Smith, Daniel Khashabi, and Hannaneh Hajishirzi. 2023.
\newblock \href {http://arxiv.org/abs/2212.10560} {Self-instruct: Aligning language models with self-generated instructions}.

\bibitem[{Wei et~al.(2022)Wei, Wang, Schuurmans, Bosma, Chi, Le, and Zhou}]{wei2022chain}
Jason Wei, Xuezhi Wang, Dale Schuurmans, Maarten Bosma, Ed~Chi, Quoc Le, and Denny Zhou. 2022.
\newblock Chain of thought prompting elicits reasoning in large language models.
\newblock \emph{arXiv preprint arXiv:2201.11903}.

\bibitem[{Wei et~al.(2023)Wei, Wang, Schuurmans, Bosma, Ichter, Xia, Chi, Le, and Zhou}]{wei2023chainofthought}
Jason Wei, Xuezhi Wang, Dale Schuurmans, Maarten Bosma, Brian Ichter, Fei Xia, Ed~Chi, Quoc Le, and Denny Zhou. 2023.
\newblock \href {http://arxiv.org/abs/2201.11903} {Chain-of-thought prompting elicits reasoning in large language models}.

\bibitem[{Wu et~al.(2023)Wu, Fei, Qu, Ji, and Chua}]{wu2023nextgpt}
Shengqiong Wu, Hao Fei, Leigang Qu, Wei Ji, and Tat-Seng Chua. 2023.
\newblock \href {http://arxiv.org/abs/2309.05519} {Next-gpt: Any-to-any multimodal llm}.

\bibitem[{Yang et~al.(2023)Yang, Li, Lin, Wang, Lin, Liu, and Wang}]{yang2023dawn}
Zhengyuan Yang, Linjie Li, Kevin Lin, Jianfeng Wang, Chung-Ching Lin, Zicheng Liu, and Lijuan Wang. 2023.
\newblock \href {http://arxiv.org/abs/2309.17421} {The dawn of lmms: Preliminary explorations with gpt-4v(ision)}.

\bibitem[{Yao et~al.(2023)Yao, Zhao, Yu, Du, Shafran, Narasimhan, and Cao}]{yao2023react}
Shunyu Yao, Jeffrey Zhao, Dian Yu, Nan Du, Izhak Shafran, Karthik Narasimhan, and Yuan Cao. 2023.
\newblock \href {http://arxiv.org/abs/2210.03629} {React: Synergizing reasoning and acting in language models}.

\bibitem[{Yin et~al.(2023{\natexlab{a}})Yin, Fu, Zhao, Li, Sun, Xu, and Chen}]{yin2023survey}
Shukang Yin, Chaoyou Fu, Sirui Zhao, Ke~Li, Xing Sun, Tong Xu, and Enhong Chen. 2023{\natexlab{a}}.
\newblock A survey on multimodal large language models.
\newblock \emph{arXiv preprint arXiv:2306.13549}.

\bibitem[{Yin et~al.(2023{\natexlab{b}})Yin, Fu, Zhao, Xu, Wang, Sui, Shen, Li, Sun, and Chen}]{yin2023woodpecker}
Shukang Yin, Chaoyou Fu, Sirui Zhao, Tong Xu, Hao Wang, Dianbo Sui, Yunhang Shen, Ke~Li, Xing Sun, and Enhong Chen. 2023{\natexlab{b}}.
\newblock Woodpecker: Hallucination correction for multimodal large language models.
\newblock \emph{arXiv preprint arXiv:2310.16045}.

\bibitem[{Zhang et~al.(2023{\natexlab{a}})Zhang, Li, and Bing}]{zhang2023videollama}
Hang Zhang, Xin Li, and Lidong Bing. 2023{\natexlab{a}}.
\newblock \href {http://arxiv.org/abs/2306.02858} {Video-llama: An instruction-tuned audio-visual language model for video understanding}.

\bibitem[{Zhang et~al.(2023{\natexlab{b}})Zhang, Liu, and Zhang}]{zhang2023extractive}
Haopeng Zhang, Xiao Liu, and Jiawei Zhang. 2023{\natexlab{b}}.
\newblock Extractive summarization via chatgpt for faithful summary generation.
\newblock \emph{arXiv preprint arXiv:2304.04193}.

\bibitem[{Zhang et~al.(2023{\natexlab{c}})Zhang, Liu, and Zhang}]{zhang2023summit}
Haopeng Zhang, Xiao Liu, and Jiawei Zhang. 2023{\natexlab{c}}.
\newblock \href {http://arxiv.org/abs/2305.14835} {Summit: Iterative text summarization via chatgpt}.

\bibitem[{Zhang(2023)}]{zhang2023graph}
Jiawei Zhang. 2023.
\newblock Graph-toolformer: To empower llms with graph reasoning ability via prompt augmented by chatgpt.
\newblock \emph{arXiv preprint arXiv:2304.11116}.

\bibitem[{Zhang et~al.(2023{\natexlab{d}})Zhang, Press, Merrill, Liu, and Smith}]{zhang2023language}
Muru Zhang, Ofir Press, William Merrill, Alisa Liu, and Noah~A. Smith. 2023{\natexlab{d}}.
\newblock \href {http://arxiv.org/abs/2305.13534} {How language model hallucinations can snowball}.

\bibitem[{Zhuang et~al.(2023)Zhuang, Yu, Wang, Sun, and Zhang}]{zhuang2023toolqa}
Yuchen Zhuang, Yue Yu, Kuan Wang, Haotian Sun, and Chao Zhang. 2023.
\newblock \href {http://arxiv.org/abs/2306.13304} {Toolqa: A dataset for llm question answering with external tools}.

\end{thebibliography}


\end{document}